# Assessing the Effects of Hyperparameters on Knowledge Graph Embedding Quality

OLIVER LLOYD

MRC Integrative Epidemiology Unit, Bristol Medical School, University of Bristol, UK. Oliver.lloyd@bristol.ac.uk

YI LIU

MRC Integrative Epidemiology Unit, Bristol Medical School, University of Bristol, UK. Yi6240.liu@bristol.ac.uk

TOM GAUNT

MRC Integrative Epidemiology Unit, Bristol Medical School, University of Bristol, UK. Tom.gaunt@bristol.ac.uk

Embedding knowledge graphs into low-dimensional spaces is a popular method for applying approaches, such as link prediction or node classification, to these databases. This embedding process is very costly in terms of both computational time and space. Part of the reason for this is the optimisation of hyperparameters, which involves repeatedly sampling, by random, guided, or brute-force selection, from a large hyperparameter space and testing the resulting embeddings for their quality. However, not all hyperparameters in this search space will be equally important. In fact, with prior knowledge of the relative importance of the hyperparameters, some could be eliminated from the search altogether without significantly impacting the overall quality of the outputted embeddings. To this end, we ran a Sobol sensitivity analysis to evaluate the effects of tuning different hyperparameters on the variance of embedding quality. This was achieved by performing thousands of embedding trials, each time measuring the quality of embeddings produced by different hyperparameter configurations. We regressed the embedding quality on those hyperparameter configurations, using this model to generate Sobol sensitivity indices for each of the hyperparameters. By evaluating the correlation between Sobol indices, we find substantial variability in the hyperparameter sensitivities between knowledge graphs, with differing dataset characteristics being the probable cause of these inconsistencies. As an additional contribution of this work, we identify several relations in the UMLS knowledge graph that may cause data leakage via inverse relations, and derive and present UMLS-43, a leakage-robust variant of that graph.

# 1 INTRODUCTION

Substantial amounts of time and energy can be spent during the tuning of knowledge graph embeddings (KGEs) due to the computationally expensive nature of this process. This expense comes not just from algorithm complexity and dataset size, but also from the sheer breadth of the hyperparameter space. For example, optimising embeddings for a medium sized graph such as FB15k-237 (1) can take upwards of 100 hours even with multiple graphical processing units (GPUs) (2). The specific GPUs used in that study were 4x NVIDIA Tesla P100-SXM. At their maximum power draw of 300 watts, total electricity consumed for that single embedding process would have been more than 30 kilowatt-hours, which is more than a median UK household will use in 100 days (3). This figure is alarmingly high, particularly when considering that the study assessed 17 methods, each analysed over 5 datasets. With the rising popularity of graph databases (4) we expect a corresponding increase in the application of embeddings, highlighting the importance of reducing the cost of the process.

A number of approaches have been proposed to improve the efficiency of embedding KGs. Speed is one such facet for this, and algorithms such as FastRP (5) and RandNE (6) are capable of producing embeddings thousands of times faster than baseline methods such as DeepWalk or Node2vec. Achieved via random projection techniques, these approaches produce equal or better quality embeddings than those baselines at a fraction of the computational cost. The process of negative sampling, in which negative examples of edges are generated, has also come under scrutiny for its inefficiencies. A non-sampling method called NS-KGE has been developed that utilises *all* non-existent edges as examples, thereby eliminating the model instability associated with negative sampling (7). This approach does come with increased cost, so mathematical derivations are leveraged to reduce the complexity of the loss function calculation down to acceptable levels. Some researchers have done away with negative sampling altogether, opting to only use positive examples in the training process for their PROCRUSTES algorithm (8). Another efficiency improvement proposed in that same work is the aligning of tuples within a batch to be of the same relation type - this facilitates parallelisation as well as simplifying computation. Parallelisation itself does not reduce computational load but does enable analyses that would be otherwise infeasible in serial processing. Several KGE libraries offer this functionality, including LibKGE (9) and DGL-KE (10). Other research has focused on the storage of embeddings. LightKG (11) stores codebooks instead of continuous vectors, which results in a 7-fold decrease in required disk space. The authors further report that codebook storage can speed up entity lookup by more than 15-fold. This is a large efficiency improvement that becomes even more relevant when the embeddings are subject to a lot of downstream analysis (as they often are).

While these advances are important, there is room for further efficiency improvement by examining the scope of the hyperparameter optimisation process. Though shown to be important in the embedding task (12), it can be very inefficient (as noted above). This is partly because it is often carried out by random or pseudo-random grid-searches, methods that may involve a lot of redundancy if there are sets of task-irrelevant hyperparameters repeatedly being tested at different values despite having little to no effect on the quality of the outputted embeddings. With some prior knowledge these unimportant variables could remain fixed, allowing a search space with reduced complexity which places higher emphasis on the tuning of task-relevant hyperparameters. As a result, the whole tuning process should be more time- and energy-efficient.

To gain understanding into the relative importance of hyperparameters, we ran a global sensitivity analysis to ascertain the effect of changes to their values on the quality of the produced embeddings. We collected the data to support this analysis by implementing tens of thousands of embedding trials using a state-of-the-art embedding library. By repeating this process for three well-known benchmark KGs, we have enabled comparison of hyperparameter sensitivities across datasets.

## 2 METHODS

### 2.1 Datasets

The datasets included in this study are FB15k-237, UMLS, and WN18RR - three popular datasets used in KG completion research. A summary of their characteristics is displayed in Table 1, and brief descriptions of their origins are found in the following paragraphs.

Table 1: Graph characteristics for the analysed datasets. Here, 'node degree' refers to the count of all adjacent edges to a given node (i.e. in-degree + out-degree). A 'component' refers to a weakly connected subgraph that is disconnected from the rest of the graph.

| Statistic | FB15k-237 | UMLS | WN18RR |
| --- | --- | --- | --- |
| node count | 14541 | 135 | 40943 |
| edge count | 310116 | 6529 | 93003 |
| edge type count | 237 | 46 | 11 |
| density | 6.19e-06 | 7.85e-03 | 5.04e-06 |
| component count | 6 | 1 | 13 |
| mean component diameter | 2.5 | 2 | 2.46 |
| mean component distance | 1.40 | 1.61 | 1.55 |
| mean component connectivity | 1 | 4 | 1.15 |
| mean node degree | 42.65 | 96.73 | 4.54 |
| median node degree | 26 | 71 | 3 |
| maximum node degree | 8642 | 382 | 521 |
| standard deviation of node degree | 127.70 | 87.44 | 8.58 |
| skewness of node degree | 34.07 | 1.84 | 26.15 |
| kurtosis of node degree | 1777.59 | 2.91 | 1095.90 |

Freebase is a knowledge graph that was initially created in 2007 by Metaweb. Google acquired the company in 2010 and shut it down a few years later, transferring its data to Wikidata in the process. The FB15k dataset (13) is a subset of 592,213 edges from Freebase, comprising 14,951 entities (hence the name) and 1,345 meta-relations that make up a selection of general facts such as the following triple: ('Jackie Chan', 'Acted in', 'Around the world in 80 days'). In a 2015 paper Toutanova and Chen highlighted the issue of inverse relations in the dataset, showing that a simple ruled-based approach could achieve state-of-the-art performance levels because many test set relations are just the inverse of those in the training set. For example, the aforementioned triple would be problematic if the following triple also existed: ('Around the world in 80 days', 'starred', 'Jackie Chan'). By removing 411 entities and including just 237 of the original relations, they derived a leakage-robust variant of the dataset, FB15k-237, which is now used frequently in KG completion research.

WordNet (14) is a large network representing the English language, with words connected if there is a conceptual link between them. For example, the 'hypernym' relation could exist from the node 'chair' to 'furniture', or the meronymic relation 'has_part' could link 'cat' and 'tail'. WN18, also introduced by Bordes et al, is a subset of WordNet that consists of 18 meta-relations and 40,943 entities. Much like FB15k, WN18 was commonly used in KG completion research until it was reported that this dataset also suffered from inverse relation test leakage (15). To replace it the authors proposed the robust subset WN18RR, which removed 7 problematic meta-relations from the graph while retaining the same entities.

The Unified Medical Language System (UMLS) (16) is a set of resources that aims to consolidate many differing biomedical lexicons to standardise and facilitate interoperability between them. In graphical format, it exists as a set of 135 high-level biomedical entities (e.g. 'vitamin', 'alga', 'lipid') connected by 46 cognate relations (e.g. 'causes', 'disrupts', 'contains'). Although used in some KG completion research (e.g. (17), (18)), UMLS is less prevalent in the field than either FB15k-237 or WN18RR. Its inclusion in this work was motivated by its nature as a smaller and much denser graph than the other two, and because it represents a more specific knowledge domain. To our knowledge UMLS has not been checked for inverse relation leakage as the other two have, so here we have performed this analysis and shared a new leakage-robust variant of the graph.

**2.2 KGE methods**

For this experiment we utilised all methods available in the knowledge graph embedding library LibKGE (9). Spanning almost a decade and a variety of architectures, these methods represent some of the most popular and impactful KGE techniques developed to date. The inclusion of earlier methods such as RESCAL is justified by the LibKGE developers because older architectures can achieve comparable performance if they are allowed usage of the updated training methods and functions that are often presented alongside newer KGE methods (19). An overview of these methods is presented in Table 2 below. Please note that throughout this paper, the term 'model' will be used only to refer to an *instantiated* version of a particular KGE method/algorithm.

Table 2: KGE methods included in the study, alongside their citation and mechanism category (as described by (2)).

| Method | Mechanism | Citation |
| --- | --- | --- |
| ComplEx | Matrix factorisation | Trouillon et al, 2016 |
| ConvE | Deep learning | Dettmers et al, 2018 |
| CP | Matrix factorisation | Lacroix, Usunier, and Obozinski, 2018 |
| Distmult | Matrix factorisation | Yang et al, 2014 |
| Relational Tucker3 | Matrix factorisation | Wang, Broscheit, and Gemulla, 2019 |
| Rescal | Matrix factorisation | Nickel, Tresp, and Kriegel, 2011 |
| RotatE | Geometric | Sun et al, 2018 |
| SimplE | Matrix factorisation | Kazemi and Poole, 2018 |
| TransE | Geometric | Bordes et al, 2013 |
| Transformer * | Deep learning | Chen et al, 2020 |
| TransH | Geometric | Wang et al, 2014 |

* Transformer is based on the 'no context' HittER method (20).

## 2.3 Data collection

Hyperparameter ranges were taken from those used in the grid-searches by Ruffinelli *et al* (19), but applied to the rest of the methods implemented in LibKGE. These ranges are presented in supplementary table 1, alongside brief descriptions of the hyperparameters' effects. Following the example of the LibKGE developers, the analysis was divided into individual jobs, each representing a different combination of dataset, KGE method, training method, and loss function. For example, the TransE algorithm applied to the UMLS dataset, training under the 1vsAll approach and using binary cross-entropy (BCE) loss, would be one such job. With these four elements fixed, we then altered the other hyperparameters in individual embedding trials. A full list of the 233 jobs is provided as supplementary table 2.

100 trials were run for each job, except for FB15k-237 jobs, on which we ran only 50 due to the dataset's size and the resulting computational cost. Hyperparameter values at each embedding attempt were chosen by Sobol sequence (21) to ensure a good spread of trial points throughout the search space. Datasets were split into training, validation, and testing sets - the former two splits were used for the learning process and the latter was used to assess embedding quality via link prediction (LP). A modified version of mean reciprocal rank (MRR) was used to measure LP performance, whereby existing triples are filtered from the ranked scores in order to prevent a model being unfairly marked down. Hits@k was a viable alternative metric, but MRR's continuous nature meant that the measurement would be more robust to the presence of false-negative edges in the entity rankings. For example, if a model scores a false-negative edge highly, this is a good indication that the embeddings are truly representative of the underlying graph. However, if that same edge pushes the target edge out of the top k rankings then the model will be marked down for this under hits@k evaluation, which would be incorrect. MRR does not entirely solve this issue, but it does lessen the impact.

## 2.4 Sensitivity analysis

For each KG, we selected the top 5% of trials by embedding quality to eliminate those that may have been poor quality due to stochastic processes. These performant trials' MRR scores were then regressed on their corresponding hyperparameter configurations using linear regression from scikit-learn (22). Then, using SALib (23), we performed a Sobol sensitivity analysis (24) on each of the regression models and recorded the output indices.

Sobol indices represent the proportion of the decomposed output variance that is attributable to a particular input, or set of inputs, of a regression model. First-order indices correspond to the variance caused by each input variable alone, second-order indices correspond to variance attributed to interactions between *pairs* of input variables, and so on. Total-order indices are simply the sum of first- and all higher-order indices for each model input. As an example, if we have a general model $Y = f(X)$ with *n* inputs and total output variance *V*, first-order sensitivities, $S_1$, can be written as follows:

$$S_1 = \{\alpha_1, \alpha_2, .., \alpha_n\} \quad (1)$$

where $\alpha_i$ is:

$$\alpha_i = E(\Delta Var(Y)|\Delta x_i)/V \quad (2)$$

with $x_i$ being the i-th input variable of X, such that:

$$\sum_{i=1}^{n} \alpha_i <= 1 \quad (3)$$

We can write second-order sensitivities, S2, as:

$$S_2 = \{\beta_{ij} \; for \; i, j \; in \; X, \; i \neq j\} \quad (4)$$

where $\beta_{ij}$ is:

$$\beta_{ij} = E(\Delta Var(Y)|\Delta x_i, \Delta x_j)/V \quad (5)$$

Third- and higher-order indices follow the same pattern but are not calculated by SALib so we will not define them here nor use them in the following equations. With that in mind, total-order sensitivities, $S_T$, can be written similarly to the first-order set:

$$S_T = \{\omega_1, \omega_2, .., \omega_n\} \quad (6)$$

But in the total-order case, $\omega_i$ is calculated as:

$$\omega_i = \alpha_i + \sum_{j=1}^{n} \beta_{ij}, \; i \neq j \quad (7)$$

Applied in this case, we can view the Sobol indices as indicators of the relative importance of different KGE method hyperparameters, i.e. those with higher indices should be prioritised in the hyperparameter tuning process.

SALib uses the Monte Carlo approach for index estimation, so to this end a total of 86,016 samples were taken from each of the OLS models' input spaces using Saltelli's extension of the Sobol sequence (25). This figure is slightly lower than the 100,000 recommended for the 20 input parameters that our regression models have (26), but due to the inclusion of several dummy variables in our inputs we considered this acceptable. The dummies themselves were set to 'one-hot' in each sample by setting the dummy with the highest sampled continuous value to 1, and its counterparts to 0.

Once sensitivity results were calculated, we estimated pairwise Pearson's correlation across the three datasets' indices to quantify the level of agreement between them. As second-order indices form a matrix, these were flattened to 1 dimension (in row major order) prior to this analysis.

## 2.5 Experimental conditions

This work was carried out using the computational facilities of the Advanced Computing Research Centre, University of Bristol - http://www.bristol.ac.uk/acrc/. The specific environment was CentOS-7 running Python 3.8.12 with PyTorch 1.7.1, accelerated with CUDA 11.4 on 4x NVIDIA GeForce RTX 2080 Ti. The software version for scikit-learn was 0.24.2 and for SALib was 1.4.5. All code and data used in the analysis are available in our GitHub repository (27).

## 3 RESULTS

### 3.1 Embedding quality

We first report, in Figure 1, on our proxy for embedding quality: LP results. Immediately noticeable is the MRR for the UMLS dataset, whose median value is consistently more than twice that of the other two KGs, regardless of method. Due to the nature of our measure for embedding quality, however, we cannot make direct comparisons between KGs for the embedding quality, only within KGs. For example, the lower-end results of the CP method on WN18RR can be said to be the poorest quality for that dataset, but we cannot definitively say that those embeddings are better/worse than any of those from FB15k-237 or UMLS. The explanation for the MRR difference is therefore that the UMLS graph just presents an easier LP problem. Because WN18RR and FB15k-237 have both been constructed to specifically remove inverse relations, our first suspicion was that this performance-boosting problem is present in UMLS. Analysis revealed three relations that should be considered problematic by Dettmers' definition (15): 'degree_of', 'precedes', and 'derivative_of'. However, these relations represented fewer than 2% of the held-out edges and as such cannot be the sole reason for the KGE methods' improved MRR on UMLS. All edges in UMLS that have a corresponding non-zero inverse edge, including those that did not cross the threshold to be considered problematic, are presented in supplementary table 4. Dataset statistics (see Table 1) may offer an additional explanation for the performance disparity. The UMLS graph is substantially denser than the other two, at 0.008 compared to 5e-6 and 6.2e-6 for WN18RR and FB15k-237 respectively. This should, on average, allow the models more contextual information with which they can make predictions. Furthermore, median degree is much higher, and skewness of degree is much lower in UMLS than the other datasets, meaning there will be far fewer situations where the model is presented with a low-context problem and that would produce a lower confidence prediction.

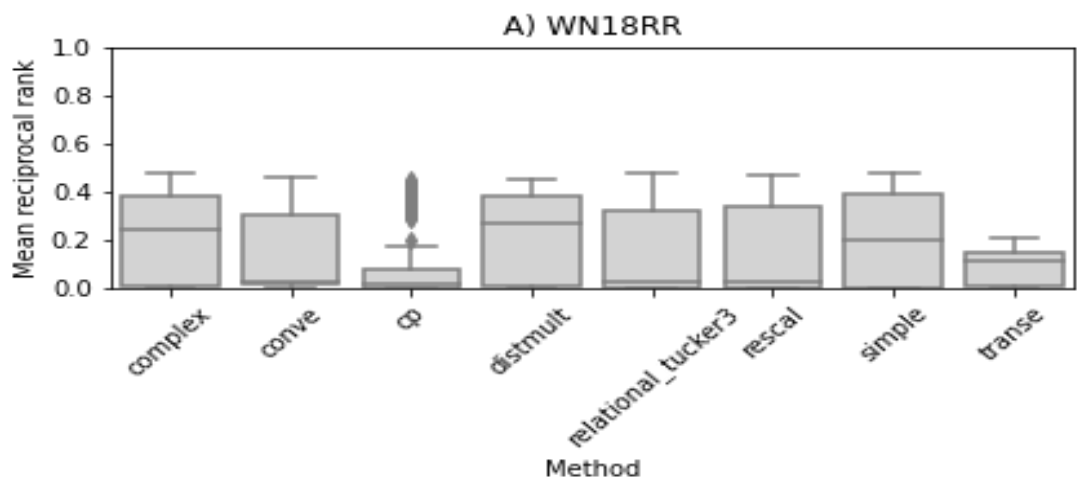

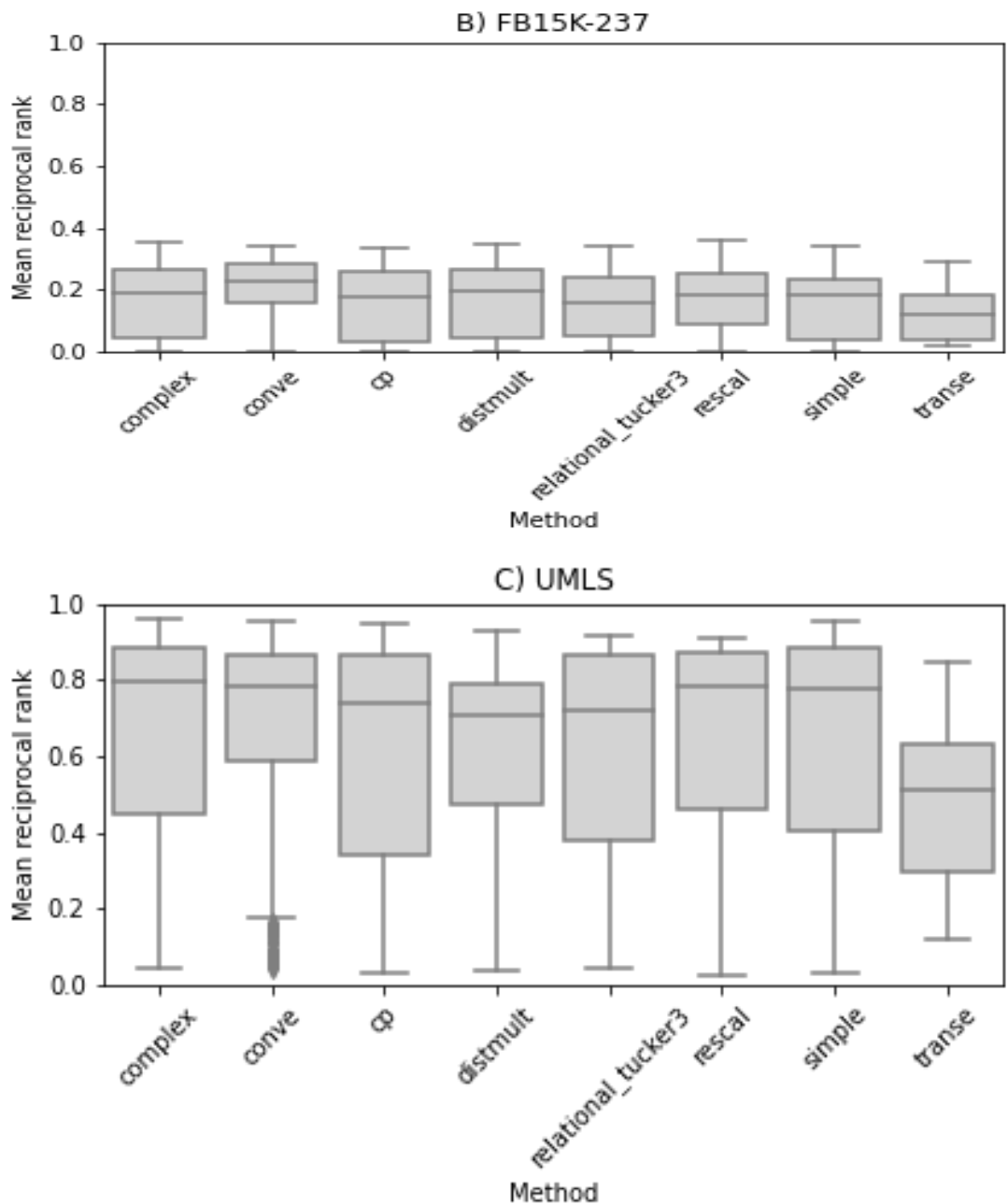

Figure 1: Box plot showing link prediction performance by method and dataset. Outliers are displayed as black diamonds.

The top results shown here do not differ much from the majority of those reported for the same three datasets (28) (29) (30), giving us confidence in the quality of our embeddings for the downstream analysis. Several of the employed KGE methods, however, were unable to be successfully run. From the selection of 233 total jobs, 36 were deemed invalid by LibKGE prior to running - 15 of these came from TransE, which the software will only run if the training method is negative sampling and the loss function is one of BCE or margin ranking. All 21 Transformer/HittER jobs were invalid because the method does not support the '_po' scoring method which was used in all other jobs. On the other hand, some jobs did not complete regardless of their validity. Both RotatE and TransH failed on all jobs for WN18RR and FB15k-237 with full computational power allocated, likely due to the size of the datasets and/or inefficiencies in the specific implementations used. TransH also failed on UMLS when using 1vsAll and KvsAll training with BCE loss. As a result of these failures, these 2 KGE methods were excluded from the UMLS sensitivity analysis in order to make the resulting indices comparable with those of the other two KGs. Overall, of the 197 valid jobs, 167 ran to completion; this corresponds to a total of 13,450 embedding trials.

### 3.2 Hyperparameter sensitivities

For each of our per-dataset sensitivity results we report: first- and total-order Sobol sensitivity indices as bar charts (Figure 2), and second-order sensitivities as network diagrams (Figure 3).

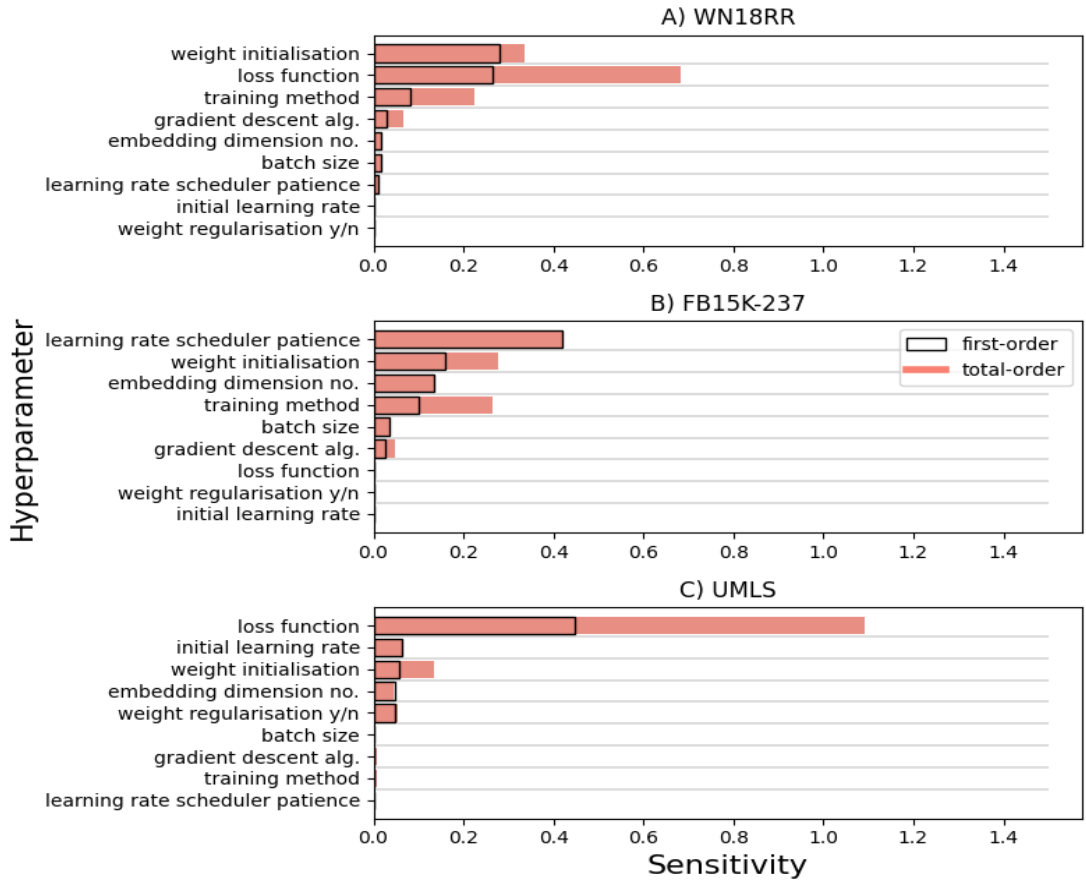

Figure 2: Bar charts showing Sobol sensitivities of the hyperparameters of knowledge graph embedding methods on three benchmark datasets. Note that dummy variables and continuous subparameters have been cumulatively grouped into their corresponding categories: training method, loss function, weight initialisation, and gradient descent algorithm. The raw data used to generate the figure are provided in supplementary table 3.

For the UMLS embeddings we see in Figure 2 that loss function dominates the relative importance of hyperparameters, both in terms of first- and total-order sensitivity. This is somewhat mirrored in WN18RR, where we see loss function have the highest total-order, despite being beaten by a slim margin on first-order sensitivity by weight initialisation methods. FB15k-237, on the other hand, seems to show no sensitivity at all to the choice of loss function. In fact, its most sensitive hyperparameter is the patience of the learning rate scheduler (ReduceLROnPlateau), a value that is almost completely unimportant when embedding the other two KGs. Another area of inconsistency is the response to the size of the embedding space. This particular hyperparameter's sensitivity is relatively high for FB15k-237, perhaps attributable to the size and complexity of that graph, but again is negligible for the other two. Surprisingly, training method achieves just middling ranks on WN18RR and FB15k-237, and registers almost no sensitivity at all on UMLS apart from some very minor higher-order sensitivity. Also unexpectedly, the initial learning rate value actually scores above 0 sensitivity in one case. This occurs on UMLS where it finds itself ranked 2nd by first-order sensitivity on this dataset, but is still a long way behind loss function in 1st place and barely ahead of the three hyperparameters below it.

A few consistencies are arguably observed in Figure 2, mainly for weight initialisation which ranks in the top 3 hyperparameters across all three datasets. However, there is still quite a large reduction in its sensitivity on UMLS, where, save for its total-order value, it does not stand out from the other non-zero hyperparameters. Choice of gradient descent algorithm, batch size, and whether to regularise weights, could all also be considered to be consistent in their sensitivities, in that they are always less than 0.1 for both first- and total-order. However, they still differ across the datasets in whether they are actually zero or not.

Overall, it is apparent that there is substantial variation in the relative hyperparameter sensitivities across the KG datasets. In fact, pairwise Pearson's correlation analysis on the ungrouped Sobol indices reveals that the highest correlation between first-order sensitivities of any two datasets is only 0.047 (WN18RR - UMLS). The other two pairs cross over into negative correlation, with WN18RR - FB15k-237 achieving a score of -0.13, and UMLS - FB15k-237 reaching the shallow depth of -0.19.

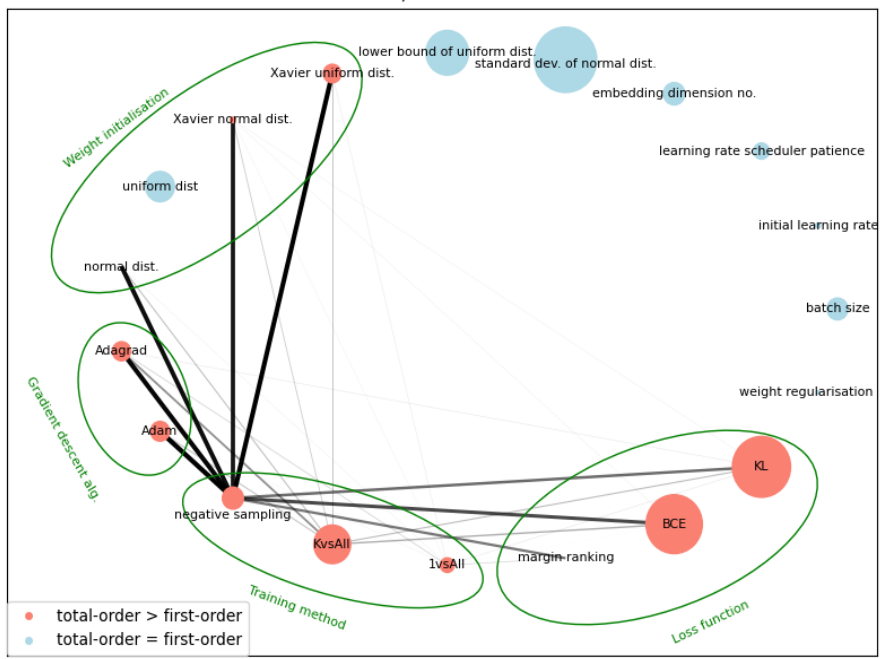
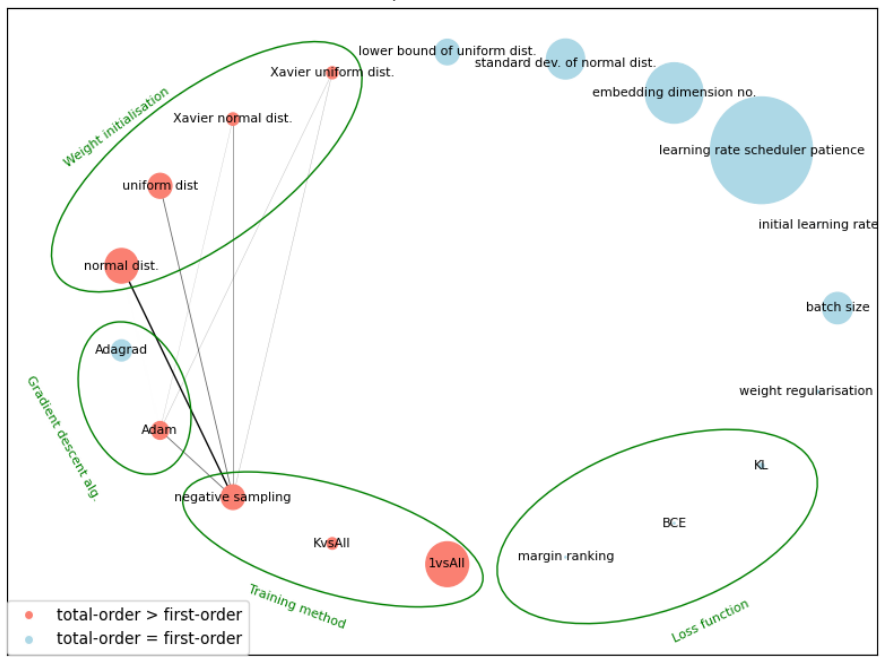

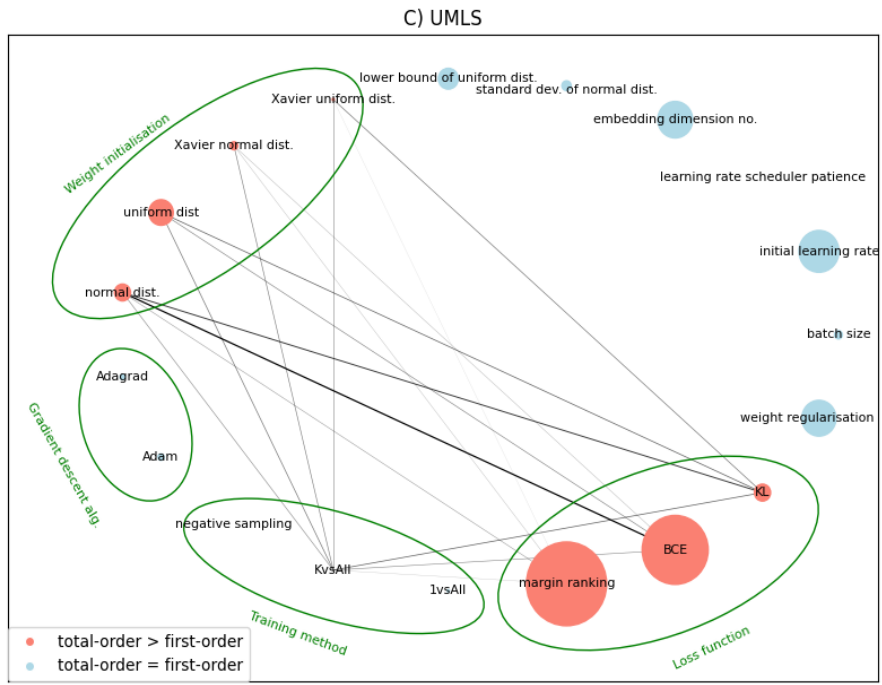

Figure 3: Network diagrams for second-order Sobol sensitivities between hyperparameters of KGE methods on three benchmark datasets. Width and opacity of edges represent the magnitude of second-order effects. Node size represents first order sensitivity. Dummy variables are not grouped in this figure, but edges between them are disallowed as they are meaningless. Dummies' categories are indicated by green bands around nodes. Raw adjacency matrices for these networks are provided in supplementary tables 5.1, 5.2, and 5.3 respectively.

In Figure 3 we see the strongest second-order sensitivities when embedding the WN18RR dataset, particularly using the 'negative sampling' training method. This hyperparameter value interacts heavily with others from all other dummy categories, making it the most central node in this network by weighted-degree. Its counterparts, KvsAll and 1vsAll, are also well connected, though the magnitude of their sensitivities is lower. For UMLS, training methods are still well connected but we see slightly heavier edges between weight initialisation and loss functions. This dataset also sees minor interactions involving gradient descent algorithm dummies, something that is mirrored by WN18RR but not by FB15k-237. Interestingly, across all three datasets we see that continuous hyperparameters (upper right nodes, clockwise from 'lower bound of uniform dist.' round to 'batch size') have negligible second-order sensitivities, meaning that such interactions are only occurring between the one-hot encoded categorical hyperparameters.

At the second-order level then, there superficially appears to be more consistency in hyperparameter sensitivity between KGs. Since our networks in Figure 3 are undirected and all contain the same nodes, we can perform pairwise Pearson's correlation of them by flattening the upper triangles of their adjacency matrices to get 1-D vectors. Doing so reveals that the correlation levels do not, in fact, improve upon those of the first-order indices. WN18RR - UMLS scores 0.078, WN18RR - FB15k-237 scores 0.063, and UMLS - FB15k-237 once more dips into negative correlation at -0.08.

## 4 DISCUSSION

Hyperparameter tuning with the goal of generating high quality/representative KG embeddings is an expensive task, in part due to the size of the hyperparameter search space. With prior understanding of the relative contributions of different hyperparameters to the variance of that quality, we could trim down the search space so that we are not spending time and energy altering inconsequential variables. Here though, we have found that Sobol sensitivity indices of KGE algorithm hyperparameters differ substantially when embedding different datasets, and this difference is observed for higher-order indices as well as the first-order. Some of the intra-dataset sensitivity results presented here might prove useful in future embedding tasks on these specific datasets, for instance the lack of importance of loss function for FB15k-237 (Figure 2.B). However, these inferences are not generalisable to external datasets.

Varying graph characteristics seem the most likely cause of the reported differences in hyperparameter sensitivities across different KGs. For example, the '1vsAll' training strategy may not work well when embedding a graph with high density (e.g. $> 0.5$), due to the fact that the method does not exclude existing edges from the negative examples it generates (31). Thanks to the density, the embedding model might then frequently be presented with the same edges as both positive and negative examples, rendering the examples useless and slowing down the learning process. If this was the case, the method would therefore be more sensitive to the choice of training method (i.e. not 1vsAll) than it would be if the graph itself was less dense. As a more concrete example, we see here that our biggest graph, FB15k-237, is also by far the most sensitive to changes to the dimensional size of the embeddings. Given the diverse statistics of the graphs analysed here (see Table 1), it is perhaps unsurprising that we see a lack of consistency in their hyperparameter sensitivities.

Of course, other effects could be influencing results in parallel, if only to a lesser degree. For instance, the differing domains of the datasets: UMLS sits in the biomedical category, WN18RR represents language, and FB15k-237 represents 'general facts'. Real-world networks may have different patterns of edge emergence depending on their domain (32). Consequently, the ability to predict held-out edges well for each of our datasets might require very different embedding configurations, which, in turn, could cause KGE hyperparameters to be of varying importance. Our choice of metric (MRR) may offer an additional explanation for the sensitivity differences. The use of mean *reciprocal* rank has been discouraged by Fuhr (33), who stated that the use of the reciprocal changes the metric from an interval to an ordinal scale, rendering a mean calculation invalid. Others have rejected this claim, however, with Sakai (34) writing in the same forum to challenge Fuhr's commandment-style assertions and highlight that the ordinal-means issue is still being debated. Regardless of controversy, MRR is still used extensively in contemporary research to measure LP performance. Finally, in any model that initialises to a random state there is the possibility of stochastic processes impacting the outcome. Our exclusion of all but the top performing trials should lessen the impact of these processes on our results, but we cannot claim that this completely alleviates the possible effects.

In conclusion, we have shown that the response of embedding quality to changes in hyperparameter values differs substantially between KGs. To properly assess the extent to which this is determined by specific dataset attributes, it will be necessary to follow up this work with a sensitivity analysis in which permutations of the KGs are created to provide a continuous spectrum of graph characteristics that can be fed, alongside hyperparameter values, into a unified regression model. We believe this may uncover important second-order interactions that would help to explain the results presented here. In this work we have also identified a set of relations in the UMLS graph that could cause data leakage through inverse relations. We therefore present the UMLS-43 graph, a derivation of UMLS that is robust to such leakage, which is available for download in the provided GitHub repository (27).

**ACKNOWLEDGMENTS**

This work was supported by the UK Medical Research Council (MRC) and carried out in the MRC Integrative Epidemiology Unit MC_UU_00011/4.

The authors would like to thank Patrick Rubin-Delanchy for his helpful comments and insights.